\documentclass{article}

\usepackage{microtype}
\usepackage{graphicx}
\usepackage{booktabs} % for professional tables

\usepackage{hyperref}

\usepackage[accepted]{icml2024}

\usepackage{amsmath}
\usepackage{amssymb}
\usepackage{mathtools}
\usepackage{amsthm}
\usepackage{subcaption}

\usepackage{multirow}

\usepackage[capitalize,noabbrev]{cleveref}

\theoremstyle{plain}

\theoremstyle{definition}

\theoremstyle{remark}

\usepackage[textsize=tiny]{todonotes}

\usepackage{caption}

\icmltitlerunning{EAGLE: Speculative Sampling Requires Rethinking Feature Uncertainty}

\begin{document}

\twocolumn[
\icmltitle{\includegraphics[width=0.06\textwidth]{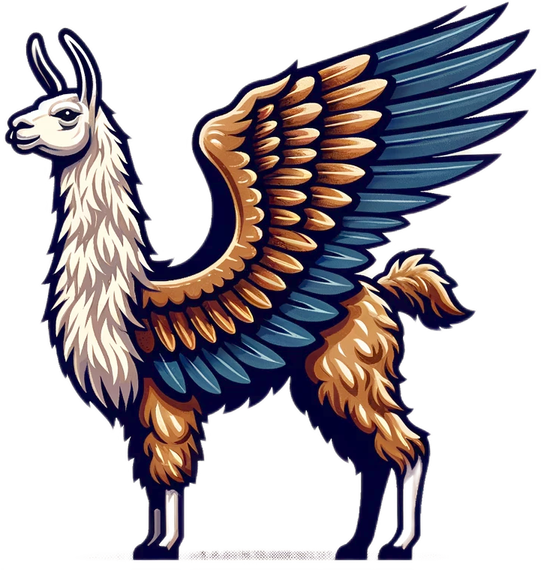} EAGLE: Speculative Sampling Requires Rethinking Feature Uncertainty}



\icmlsetsymbol{equal}{*}

\begin{icmlauthorlist}
\textbf{Yuhui Li}$^\spadesuit$
\quad\textbf{Fangyun Wei}$^\ddag$
\quad\textbf{Chao Zhang}$^\spadesuit$
\quad\textbf{Hongyang Zhang}$^\clubsuit$$^\dag$
\\
$^\spadesuit$Peking University\quad $^\ddag$Microsoft Research\quad $^\clubsuit$University of Waterloo\quad $^\dag$Vector Institute\\
\texttt{hongyang.zhang@uwaterloo.ca}
\\
\url{https://github.com/SafeAILab/EAGLE}
\end{icmlauthorlist}

% \begin{icmlauthorlist}
% \icmlauthor{Yuhui Li}{yyy}
% \icmlauthor{Fangyun Wei}{comp}
% \icmlauthor{Chao Zhang}{yyy}
% \icmlauthor{Hongyang Zhang}{sch,vec}
% \\
% \url{https://github.com/SafeAILab/EAGLE}
% \end{icmlauthorlist}

\icmlcorrespondingauthor{Hongyang Zhang}{hongyang.zhang@uwaterloo.ca}

\icmlkeywords{Machine Learning, ICML}

{%
\renewcommand\twocolumn[1][]{#1}%
\begin{center}
    \centering
    \captionsetup{type=figure}
    \includegraphics[width=1\textwidth]{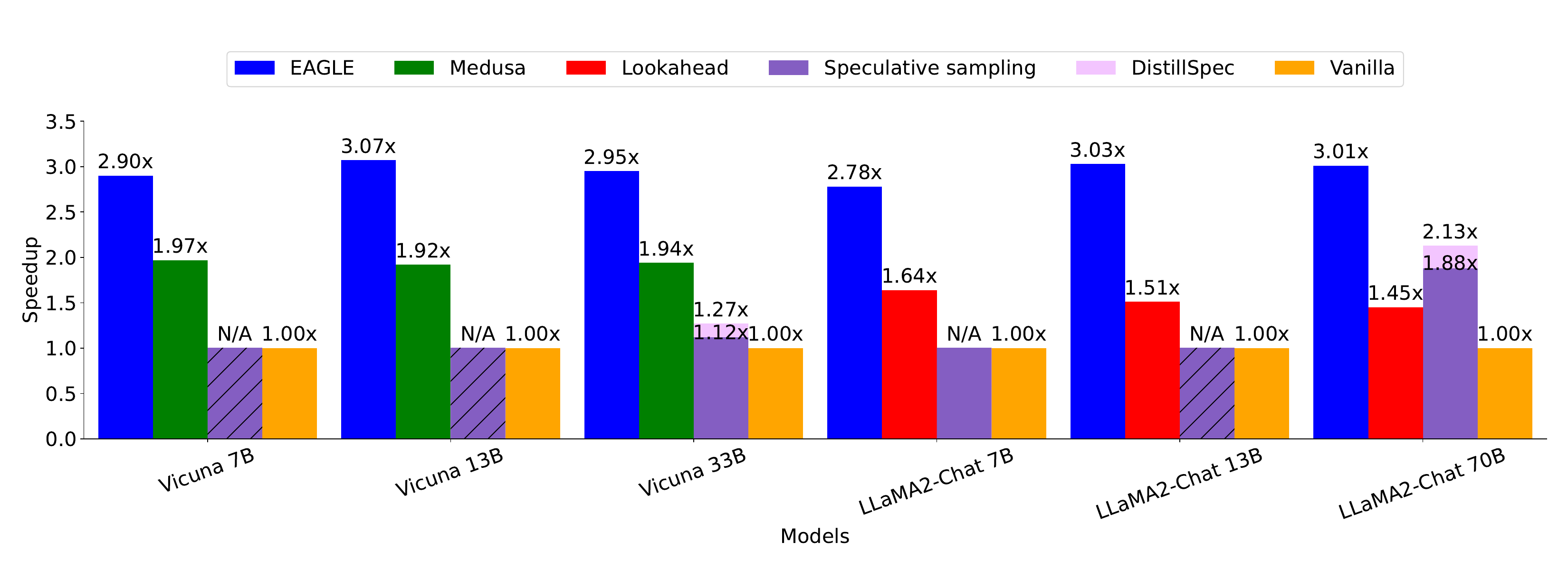}
    \vspace{-0.6cm}
    \captionof{figure}{Speedup ratio of Vicuna and LLaMA2-Chat inference latency on the MT-bench for greedy (temperature=0) settings. Speedup ratio of Medusa and Lookahead are copied from their original technical reports. With speculative sampling, there is a lack of suitable draft models to accelerate the 7B model. Employing a 7B model as the draft model for a 13B model results in slow speeds due to the high overhead of the 7B model, rendering it less efficient than vanilla autoregressive decoding. These scenarios are marked as N/A. \emph{In this paper, we only compare with speculative sampling based methods that do not need to finetune the backbone models, ensuring the output text distribution remains constant.}}
    \label{mt-t0}
\end{center}%
}

\vskip 0.3in
]


%\printAffiliationsAndNotice{} % otherwise use the standard text.

\begin{abstract}
Autoregressive decoding makes the inference of Large Language Models (LLMs) time-consuming. In this paper, we reconsider speculative sampling and derive two key observations. Firstly, autoregression at the feature (second-to-top-layer) level is more straightforward than at the token level. Secondly, the inherent uncertainty in feature (second-to-top-layer) level autoregression constrains its performance. Based on these insights, we introduce EAGLE (Extrapolation Algorithm for Greater Language-model Efficiency), a simple yet highly efficient speculative sampling framework. By incorporating a token sequence advanced by one time step, EAGLE effectively resolves the uncertainty, enabling precise second-to-top-layer feature prediction with minimal overhead. We conducted comprehensive evaluations of EAGLE, including all models from the Vicuna and LLaMA2-Chat series, the MoE model Mixtral 8x7B Instruct, and tasks in dialogue, code generation, mathematical reasoning, and instruction following. For LLaMA2-Chat 70B, EAGLE achieved a latency speedup ratio of \textbf{2.7x-3.5x}, doubled throughput, while maintaining the distribution of the generated text.
\end{abstract}

\section{Introduction}

Autoregressive decoding, the de facto standard for large language models (LLMs), generates tokens sequentially, leading to slow and costly generation. 
Speculative sampling \cite{leviathan2023fast,chen2023accelerating} based methods address this by dividing the process into a low-cost draft stage and a \emph{parallelized} verification stage over the drafted tokens, allowing for multiple tokens to be validated in a single LLM pass. These approaches accelerate generation by producing multiple tokens per pass. More importantly, the verification stage ensures that the text distribution aligns precisely with the decoding results of the original LLM, maintaining the integrity of the generated content.

Applying speculative sampling hinges on finding a draft model that mirrors the original LLM's functionality but with reduced latency, often involving a lower-parameter version from the same LLM series. For instance, in the LLaMA2 \cite{touvron2023llama} series which includes models with 7B, 13B, and 70B parameters, using the 7B model as a draft model of the 70B model is valid, while finding a suitable draft model for the smallest 7B variant is tricky. An alternative could be to use TinyLLaMA \cite{zhang2024tinyllama}, but it is not feasible for instruct-tuned models due to the inconsistency in instruction templates between LLaMA2-Chat and TinyLLaMA-Chat. Despite the 7B model's potential as a draft model, its high overhead diminishes acceleration gains.
Training a new, appropriately sized draft model specifically for speculative sampling is not an ideal solution either due to the high cost: 
TinyLLaMA is trained on 3,000B tokens, whereas EAGLE is trained on 2-4B tokens.

The key to enhancing acceleration in speculative sampling lies in reducing the time overhead and improving the acceptance rate of the draft by the original LLM~\cite{chen2023cascade,xia2023speculative,santilli-etal-2023-accelerating}. Numerous approaches focus on reducing the overhead of the drafting phase. Lookahead \cite{fu2023lookahead} employs n-gram and Jacobi iteration, while Medusa \cite{medusa} utilizes a set of MLPs that predict tokens based on the second-to-top-layer feature of the original LLM. These strategies significantly decrease the latency in generating drafts, leading to improved acceleration. However, their effectiveness is limited by the lower accuracy of the resulting drafts, with Medusa achieving an accuracy of about 0.6, and Lookahead even lower. In contrast, our method attains an accuracy of approximately 0.8.

\begin{figure}[t]
\begin{center}
\centerline{\includegraphics[width=\columnwidth]{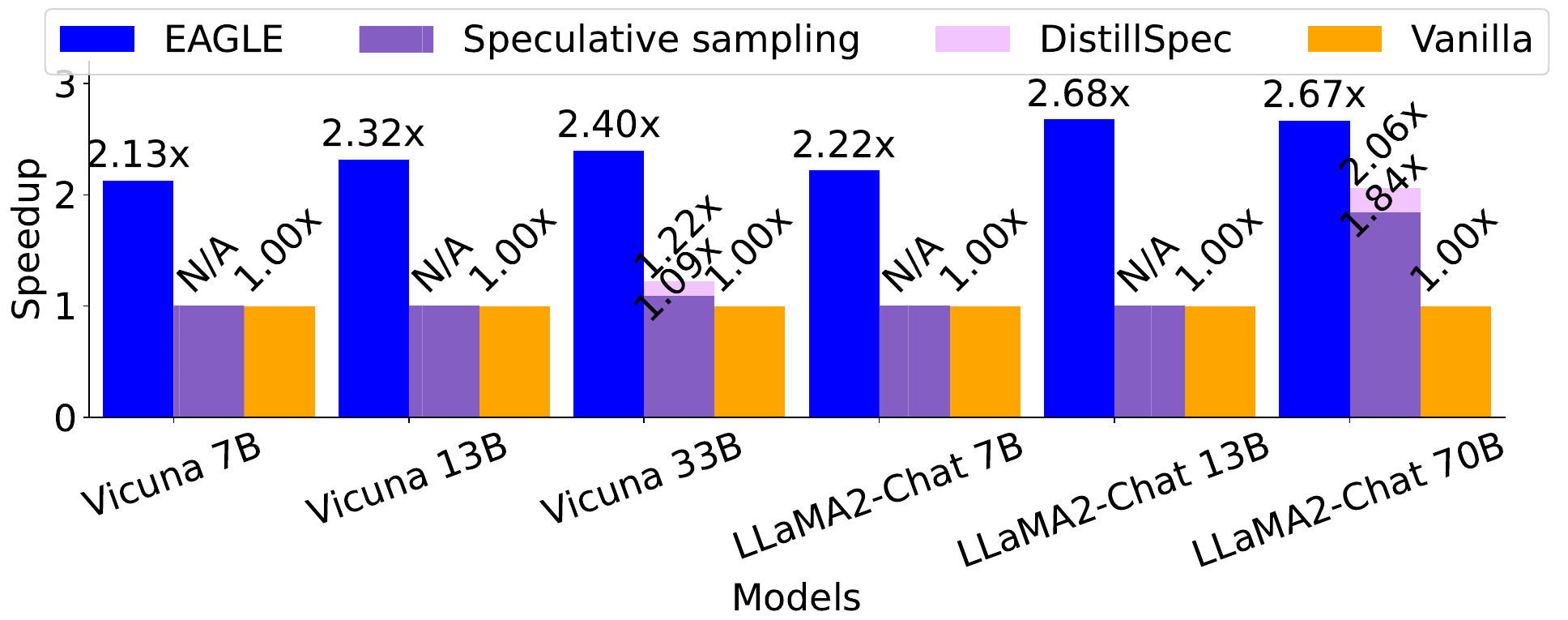}}
\caption{Speedup ratio on the MT-bench for non-greedy (temperature=1) settings. Lookahead is confined to greedy decoding, and the non-greedy generation of Medusa does not guarantee lossless performance. Therefore, EAGLE is not compared with these methods.}
\label{mt-t1}
\end{center}
\end{figure}

To overcome these limitations, we introduce EAGLE (Extrapolation Algorithm for Greater Language-model Efficiency), an efficient speculative sampling method, grounded in the following two observations.

\textbf{Firstly, autoregression at the feature level is simpler than at the token level.}  In this paper, ``features'' refer to the second-to-top-layer features of the original LLM, located before the LM head.
Compared to token sequences, which are simple transformations of natural language, feature sequences exhibit more regularity. Autoregressively processing at the feature level and then deriving tokens using the LM head of the original LLM yields better results than directly autoregressively predicting tokens. As illustrated in Figure \ref{fig:ob12}, autoregressively predicting features yields better performance, demonstrated by a higher speedup ratio of 1.9x compared to 1.5x. 

\begin{figure}
    \centering
    \includegraphics[width=\linewidth]{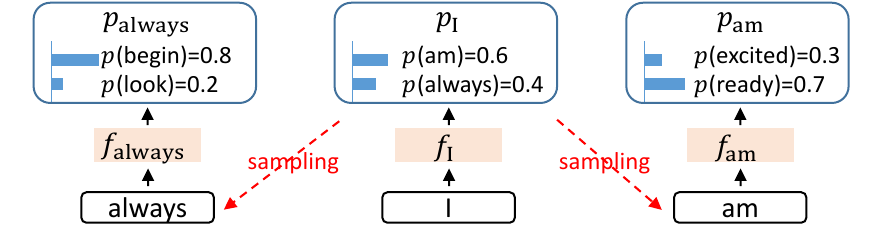}
    \caption{Uncertainty in feature sequences. The next feature following $f_I$ is contingent on the sampling outcome and cannot be determined solely based on $f_I$, where both ``always'' and ``am'' are possible to follow the token ``I'' and lead to two branches.}
    \label{fig:random}
\end{figure}

\begin{figure}[h]
    \centering
    \includegraphics[width=\linewidth]{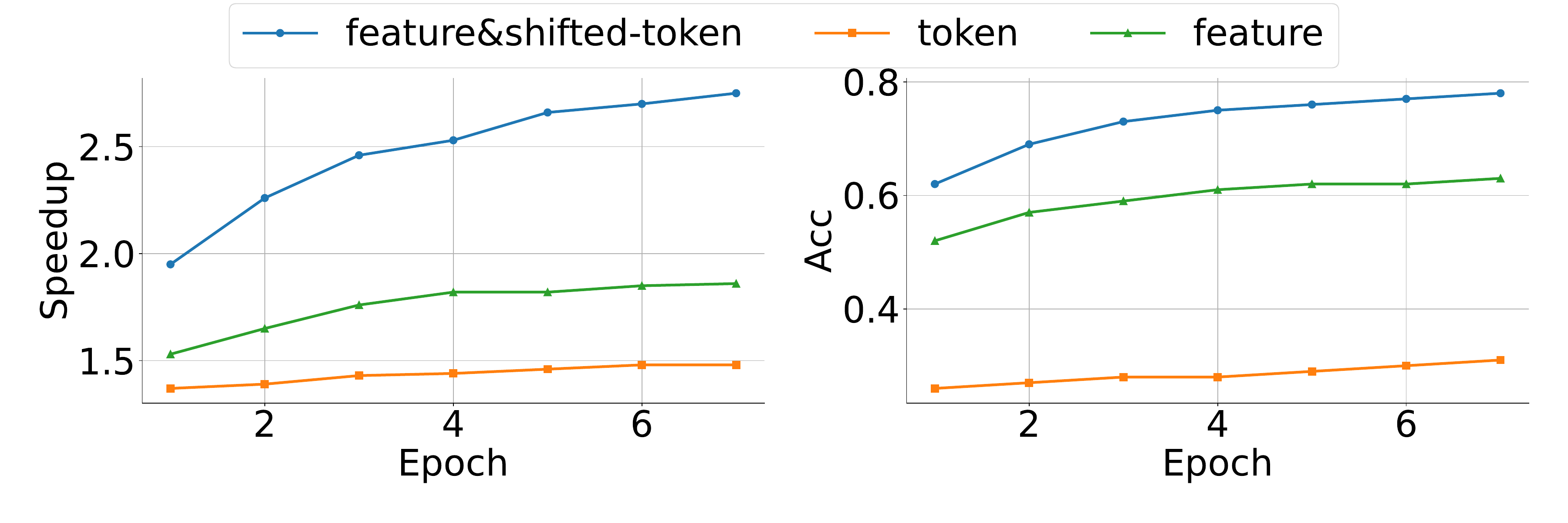}
    \caption{Accuracy and speedup ratio of draft models based on tokens, features and feature\&shifted-token at temperature=0, tested on MT-bench with Vicuna 7B as the original LLM. Feature\&shifted-token refers to using a feature sequence and a token sequence advanced by one time step as inputs.}
    \label{fig:ob12}
\end{figure}

\textbf{Secondly, the uncertainty inherent in the sampling process significantly constrains the performance of predicting the next feature.}
In text generation, the target LLM predicts the distribution of tokens and samples accordingly, introducing randomness. Features, being high-dimensional and continuous, cannot be treated similarly. As depicted in Figure \ref{fig:random}, sampling different tokens like ``am'' or ``always'' leads to distinct feature sequences, introducing ambiguity into the feature-level autoregression. Medusa faces a similar issue in predicting spaced tokens, where it is uncertain whether the true target for the input $f_I$ should be $p_{am}$ or $p_{always}$. To address this issue, EAGLE inputs the token sequence from one time step ahead, which includes the sampling outcomes, into the draft model. In the example illustrated in Figure \ref{fig:random}, this involves predicting $f_{always}$ based on $f_{I}$ and $t_{always}$, and predicting $f_{am}$ based on $f_{I}$ and $t_{am}$. As illustrated in Figure \ref{fig:ob12}, by addressing the uncertainty, the speedup ratio further increases from 1.9x to 2.8x.

We conducted experiments across dialogue, code generation, mathematical reasoning, and instruction following tasks using the MT-bench, HumanEval, GSM8K, and Alpaca datasets, respectively. Tested LLMs included all models from the Vicuna and LLaMA2-Chat series, along with Mixtral 8x7B Instruct. For LLaMA2-Chat 70B, EAGLE achieved a speedup ratio of \textbf{2.7x-3.5x}, doubled throughput, and theoretically guaranteed the preservation of the generated text's distribution.
Figure \ref{mt-t0} and \ref{mt-t1} illustrates the performance of EAGLE on the MT-bench \cite{zheng2023judging}, a highly realistic benchmark simulating actual applications and real-world scenarios, including multi-turn instructions akin to dialogues with ChatGPT. We have chosen to utilize this benchmark as it has been employed by the current state-of-the-art, including Lookahead and Medusa, to demonstrate their speedup ratios. This choice facilitates a fair and direct comparison between our approach and these methods. Compared to the recently proposed speculative sampling-based frameworks, Lookahead and Medusa, EAGLE achieves \textbf{1.7x-2.1x} and \textbf{1.5x-1.6x} speedups, respectively. 
EAGLE operates in parallel with other acceleration or throughput-improving methods, such as quantization, compilation, etc. Combining EAGLE with these techniques could further reduce the operational costs of LLM systems. For example, with gpt-fast \cite{gptfast}, EAGLE accelerates LLaMA2-Chat 7B decoding to 160.4 tokens/s on a single RTX 3090 GPU.

EAGLE boasts low training costs. For the LLaMA2-Chat 70B model, EAGLE trains a decoder layer with fewer than 1B parameters using no more than 70k dialogues from the ShareGPT dataset. The training is completed in 1-2 days on 4x A100 (40G) GPUs. The training of EAGLE on 7B, 13B and 33B models can even be conducted on a RTX 3090 node in 1-2 days. In practical applications, EAGLE requires only a single training session to provide acceleration for each query. As the number of queries increases, the amortized training cost of EAGLE becomes negligible.

Beyond performance, EAGLE offers additional advantages:
\begin{itemize}
    \item \textbf{Generality:} EAGLE is applicable to any autoregressive LLMs (at least in principle). We have applied EAGLE to LLaMA2-Chat (7B, 13B, 70B), Vicuna (7B, 13B, 33B) and Mixtral 8x7B Instruct in a zero-shot way on the MT-bench, GSM8K, HumanEval and alpaca datasets.
    EAGLE adheres to the commonly used zero-shot/few-shot settings within the LLM community. All experiments employ the same weights, trained exclusively on the ShareGPT dataset, without any additional training on the evaluation datasets. The method adds only a lightweight plug-in (a single transformer decoder layer) to the LLM, which can be easily deployed in a production environment. 
    \item \textbf{Reliability:} EAGLE does not involve any fine-tuning of the original LLM, and the preservation of the output distribution by EAGLE is theoretically guaranteed for both the greedy and non-greedy settings. 
    This is in sharp contrast to Lookahead and Medusa which either focus solely on greedy settings or do not guarantee the preservation of distribution in these settings.
 
\end{itemize}

\section{Preliminaries}

\textbf{Notations.} In this paper, ``target LLM'' denotes the LLM intended for acceleration, while ``draft model'' refers to the model used for draft generation. ``Feature'' generally signifies the second-to-top-layer feature of a LLM, the hidden state before the LM head. Tokens are denoted by lowercase $t$, their embeddings by $e$, features by $f$, and distributions by $p$. Sequences are represented in uppercase, for example, $T_{i:j}$ for $(t_i,t_{i+1},\ldots,t_{j})$. In a LLM, input $T_{1:j}$ is transformed into embeddings $E_{1:j}$ through the embedding layer, then to features $F_{1:j}$, and the LM Head maps $f_{j}$ to a distribution $p_{j+1}=\text{LM\_Head}(f_{j})$, sampling the next token $t_{j+1}$. Vanilla autoregression at the token level is described by $T_{1:j}\rightarrow E_{1:j}\rightarrow f_j\rightarrow p_{j+1}\rightarrow t_{j+1}$ for any integer $j\ge 1$.

\textbf{Speculative sampling.} 
Speculative sampling operates through draft and verification phases, with the drafting phase using a smaller model to generate $\gamma$ tokens $\hat{T}_{j+1:j+\gamma}$ and their distributions $\hat{P}_{j+1:j+\gamma}$. In the verification phase, a single forward pass of the target LLM yields the probabilities $P_{j+1:j+\gamma}$. Tokens are then sequentially evaluated, with a token $\hat{t}_{j+i}$ having an acceptance probability $\min(1,p_{j+i}(\hat{t}_{j+i})/\hat{p}_{j+i}(\hat{t}_{j+i}))$. Upon the rejection of a token $\hat{t}_{j+i}$, all subsequent tokens are discarded, and this token is resampled based on a distribution $\text{norm}(\max(0,p_{j+i}-\hat{p}_{j+i}))$. As proven in Appendix A.1 of speculative sampling \cite{leviathan2023fast}, this method equates to sampling directly from the target LLM. EAGLE adopts this method, ensuring that \textbf{the distribution of the generated text remains unchanged for both the greedy and non-greedy settings}.

\section{EAGLE}

EAGLE, aligning with other speculative sampling-based methods, incorporates both a drafting phase and a verification phase.

\subsection{Drafting phase}
\label{me:draft}

\begin{figure}
\begin{center}
\centerline{\includegraphics[width=\linewidth]{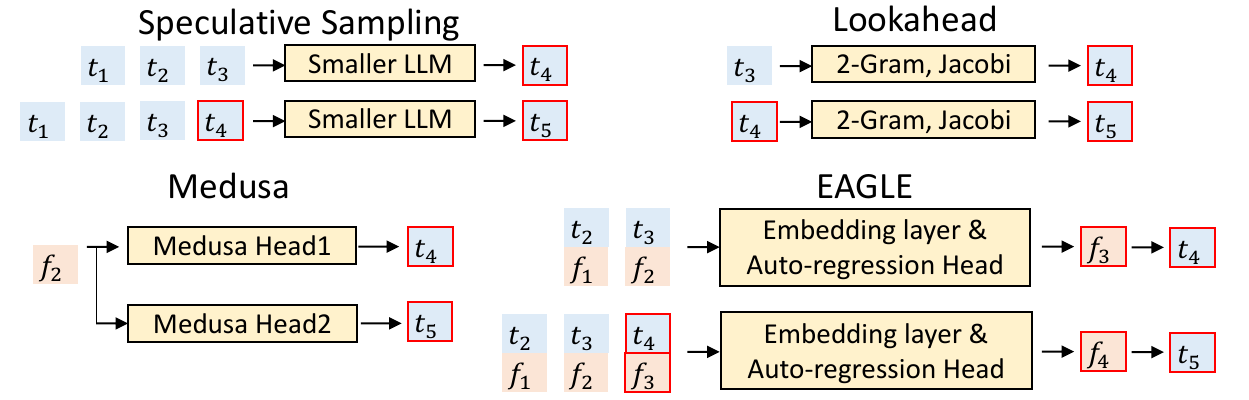}}
\caption{A comparison of the methods for drafting the fourth and fifth tokens, $t_4$ and $t_5$. $t$ (represented by \textcolor{blue}{blue} blocks) denotes tokens, and $f$ (\textcolor{orange}{orange} blocks) signifies the features, with subscripts indicating their positions in the sequence. The \textcolor{red}{red} border indicates the predictions of the draft model. For simplicity, the $n$ in the $n$-gram for Lookahead, as shown in the figure, has been set to 2.}
\label{comparation}
\end{center}
\end{figure}

The primary distinction between EAGLE and other methods lies predominantly in the drafting phase.
Figure \ref{comparation} illustrates a schematic of the drafting phase for different methods. Speculative sampling \cite{leviathan2023fast,chen2023accelerating} and Lookahead \cite{fu2023lookahead} predict tokens based on tokens. Medusa~\cite{medusa} independently predicts $t_4$ and $t_5$ using the feature $f_2$ from the target LLM. EAGLE predicts $f_3$ using the feature sequence $(f_1,f_2)$ and the token sequence $(t_2,t_3)$, advanced by one time step. From $p_4=\text{LM Head}(f_3)$, $t_4$ is sampled. Subsequently, $f_3$ and $t_4$ are concatenated into the input sequence to predict the next feature $f_4$ and sample the subsequent token $t_5$.

% Frameworks employing the draft-verification principle invariably require a drafting phase, yet EAGLE's approach to drafting is distinct from speculative sampling, Medusa, Lookahead, and similar methodologies, as illustrated in Figure \ref{comparation}. Speculative Sampling and Lookahead predict tokens based on tokens, Medusa predicts tokens based on features, whereas EAGLE uniquely predicts features based on both features and tokens.

\begin{figure}[t]
\begin{center}
\centerline{\includegraphics[width=\linewidth]{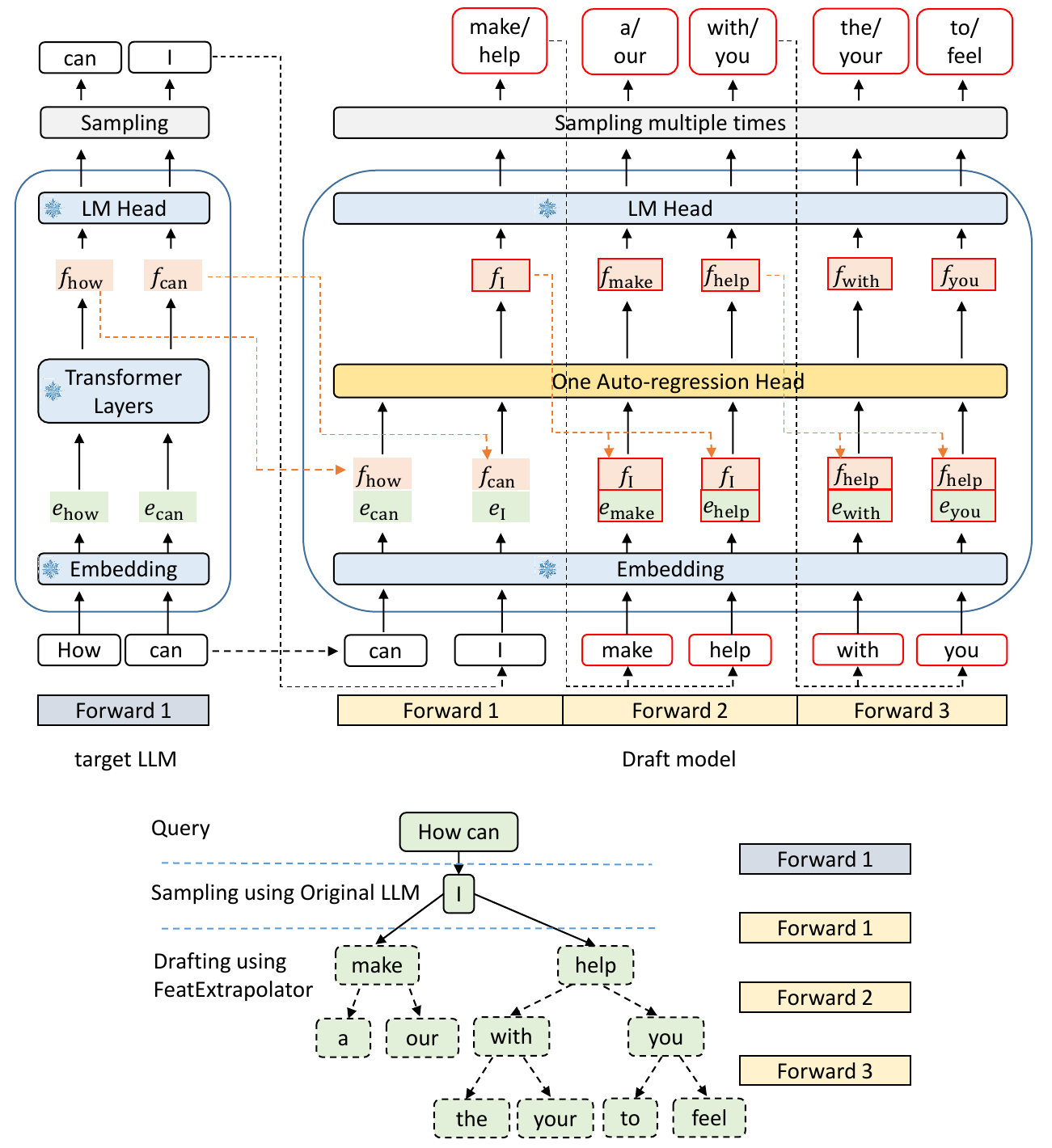}}
\caption{Pipeline of EAGLE. The upper section illustrates the computational process, while the lower section displays the corresponding generation results for each step. In the upper section, \textcolor{green}{green} blocks represent token embeddings, \textcolor{orange}{orange} blocks represent features, \textcolor{red}{red} boxes indicate the predictions of the draft model, and \textcolor{blue}{blue} modules with snowflake icons represent the use of target LLM parameters, which are not subject to training.}
\label{pipline}
\end{center}
\end{figure}

As illustrated in Figure \ref{pipline}, EAGLE's draft model comprises three modules: the Embedding layer, LM Head, and Autoregression Head. The Embedding layer and LM Head employ the parameters of the target LLM and do not necessitate additional training. The draft model takes as input a feature sequence of shape (bs, seq\_len, hidden\_dim) and an advanced token sequence of shape (bs, seq\_len). It then converts the token sequence into a token embedding sequence of shape (bs, seq\_len, hidden\_dim), and concatenates it to form a fused sequence of shape (bs, seq\_len, 2$\times$hidden\_dim). The Autoregression Head consisting of an FC layer and a decoder layer.
The FC layer reduces the dimensionality of the fused sequence to (bs, seq\_len, hidden\_dim) and then we utilize the decoder layer to predict the next feature. The LM Head calculates the distribution based on the feature, from which the next token is sampled. Finally, the predicted feature and the sampled token are concatenated into the input, facilitating the continuation of the autoregressive process. 
EAGLE creates a tree-structured draft using tree attention, generating a draft tree with depth $m$ and more than $m$ tokens through $m$ forward passes. For instance, as shown in Figure \ref{pipline}, EAGLE drafts a 10-token tree with just 3 forward passes.
The actual tree structure employed by EAGLE is detailed in Appendix \ref{ap:tree}.

\subsection{Training of the draft models}

Predicting the next feature constitutes a regression task, for which we employ Smooth L1 loss (see Figure \ref{comparation} EAGLE):
\begin{equation*}
    L_{reg} = \text{Smooth L1}(f_{i+1}, \text{Draft\_Model}(T_{2:i+1}, F_{1:i})).
\end{equation*} 
Predicting features is an intermediary objective of the draft model, with the ultimate goal being the prediction of tokens to generate a sequence of tokens. Consequently, we also employ classification loss to directly optimize towards this final objective:
\begin{gather*}
    {p}_{i+2}=\text{Softmax}(\text{LM\_Head}({f}_{i+1})), \\
    \hat{p}_{i+2}=\text{Softmax}(\text{LM\_Head}(\hat{f}_{i+1})), \\
    L_{cls} = \text{Cross\_Entropy}({p}_{i+2},\hat{p}_{i+2}).
\end{gather*}
By integrating regression loss and classification loss, we train the Autoregression Head using the combined loss function $L=L_{reg}+w_{cls}L_{cls}$.
Typically, the classification loss is an order of magnitude larger than the regression loss in numerical terms. Consequently, we set $w_{cls}$ to 0.1.

EAGLE's Autoregression Head is ideally trained with autoregressively generated text from the target LLM, yet this approach is costly. Fortunately, EAGLE exhibits low sensitivity to training data (ablation study in Section \ref{sec:abadata}). Instead of employing text generated by the target LLM, we utilize a fixed dataset, substantially reducing the overhead. 
During the drafting phase, EAGLE autoregressively processes features. Inaccuracies in features can lead to error accumulation. To mitigate this issue, we employ data augmentation by adding random noise sampled from a uniform distribution $\mathcal{U}(-0.1, 0.1)$ to features of the target LLM during training~\cite{jain2023neftune}.

\subsection{Verification phase}
Employing tree attention, the target LLM computes the probability of each token in the tree-structured draft through a single forward pass. At every node of the draft tree, we recursively apply speculative sampling algorithms to sample or adjust the distribution (details in Appendix \ref{ap:mss}), consistent with SpecInfer \cite{miao2023specinfer}, ensuring that the distribution of the output text aligns with that of the target LLM. Concurrently, we document accepted tokens and their features for use in the next drafting phase.

\label{me:ver}

\section{Experiments}

\textbf{Models and tasks.} We conducted experiments on Vicuna models (7B, 13B, 33B), LLaMA2-chat models (7B, 13B, 70B), and Mixtral 8x7B Instruct, encompassing the common sizes of current mainstream LLMs.
We evaluated EAGLE across multiple tasks including multi-turn dialogue, code generation, mathematical reasoning, and instruction following, employing the MT-bench \cite{zheng2023judging}, HumanEval \cite{chen2021evaluating}, GSM8K \cite{cobbe2021training}, and Alpaca \cite{alpaca} datasets, respectively.
Speculative sampling \cite{leviathan2023fast} conducted experiments with a batch size of 1, a setting subsequently adopted by other works such as DistillSpec \cite{zhou2023distillspec} and BiLD \cite{kim2023speculative}. Similarly, the majority of our experiments also adopted this setting. Experiments with a batch size greater than 1 are presented in Section \ref{sec:bsne1}.

\textbf{Metrics.}
Like other speculative sampling-based methods, EAGLE primarily focuses on latency rather than throughput. We assess acceleration effects using the following metrics:
\begin{itemize}
    \item Walltime speedup ratio: The actual test speedup ratio relative to vanilla autoregressive decoding.
    \item Average acceptance length $\tau$: The average number of tokens accepted per forward pass of the target LLM.
    \item Acceptance rate $\alpha$: The ratio of accepted to generated tokens during drafting, gauges draft accuracy. It's less applicable for tree drafts due to multiple tokens sampled per location with only one accepted. Hence, when measuring this metric, \textbf{we utilize chain drafts without tree attention}, aligning with speculative sampling and DistillSpec. EAGLE's draft model inputs feature and token sequences. Autoregressive feature processing can propagate errors, so we measure the acceptance rate as $n\text{-}\alpha$, considering $n$ features predicted by the draft model, potentially with inaccuracies.
\end{itemize}

Acceleration of EAGLE theoretically guarantees the preservation of the target LLMs' output distribution. Consequently, evaluating the quality of EAGLE's generated results is both unnecessary and meaningless.

\textbf{Training.}
We fixed the target LLMs. EAGLE was trained on the ShareGPT dataset, utilizing 68,000 dialogue iterations with a learning rate set at 3e-5. We employed the AdamW optimizer with beta values $(\beta_1, \beta_2)$ set to (0.9, 0.95) and implemented gradient clipping of 0.5. 
The trainable parameters of EAGLE corresponding to the 7B, 13B, 33B, and 70B models are 0.24B, 0.37B, 0.56B, and 0.99B, respectively. The trainable parameters of EAGLE for MoE model Mixtral 8x7B is 0.28B. EAGLE is characterized by its low training cost; the Autoregression Head is trainable within 1-2 days on an A100 40G server for the 70B models.

\subsection{Effectiveness}

% \begin{figure}[t]
% \begin{center}
% \centerline{\includegraphics[width=\linewidth]{./figs/task.pdf}}
% \caption{Speedup ratios of EAGLE across various tasks. }
% \label{task}
% \end{center}
% \vskip -0.2in
% \end{figure}

% Figure \ref{mt-t0} and \ref{mt-t1} illustrate the speedup ratios of EAGLE compared to other speculative sampling methods on the MT-bench. On the Vicuna 13B and LLaMA2-Chat 13B, 70B models, with a temperature setting of 0, EAGLE achieves a speed that is 3x faster than vanilla autoregressive decoding, 2x faster than Lookahead, and 1.6x faster than Medusa. At a temperature setting of 1, the speedup ratio of EAGLE is slightly lower. As shown in Table \ref{tab:other}, EAGLE's acceleration on the Alpaca dataset is similar to that on MT-bench. On the GSM8K and HumanEval datasets, EAGLE demonstrates enhanced acceleration, achieving maximum speedup ratios of 3.76x and 3.25x respectively.

Figures \ref{mt-t0} and \ref{mt-t1}, along with Table \ref{tab:other}, display the speedup ratios of EAGLE. EAGLE demonstrates better acceleration at temperature=0 compared to temperature=1. For instance, for LLaMA2-Chat 13B at temperature=0, the speedup ratios range from 3.01x to 3.76x, while at temperature=1, they range from 2.66x to 2.89x. 
In code generation tasks (HumanEval), EAGLE achieves its best acceleration performance. This is attributed to the prevalence of fixed templates in code, making it easier to generate drafts for these templates. Compared to recently introduced speculative sampling-based methods, Lookahead and Medusa, EAGLE is faster by 1.70x-2.08x and 1.47x-1.60x, respectively. 
Employing speculative sampling in the Vicuna and LLaMA2-Chat series is challenging. For the 7B model, there is no suitable draft model. For other sizes, using the 7B model as the draft model, we iterated through draft lengths from 2 to 10 and reported the highest speedup ratio. For the 13B model, we observed no improvement in speed. For the 33B and 70B models, the speedup ratios were 1.12x and 1.88x, respectively. For DistillSpec, to ensure fairness, we used the same training data as EAGLE. Additionally, the divergence function employed follows the FKL as detailed in Appendix A.1 of the DistillSpec paper.
While distillation slightly improved the speedup ratio, the limited enhancement is because distillation aims to increase the draft model's acceptance rate, while the bottleneck for speculative sampling performance lies in the high overhead of the draft model.

Tables \ref{tab:other} and \ref{tab:alpha} indicate that in EAGLE, the target LLM generates 3.2-4.5 tokens per forward pass, surpassing vanilla decoding which produces only one token per forward pass, thereby significantly increasing generation speed. 
As shown in Figure \ref{tab:alpha} and Appendix \ref{apex}, the acceptance rate for completely accurate feature sequences, $0\text{-}\alpha$, significantly exceeds that for sequences with a single erroneous feature, $1\text{-}\alpha$, indicating the impact of feature errors on draft model performance. Yet, the slight variation between $1\text{-}\alpha$ to $4\text{-}\alpha$ underscores EAGLE's robustness to feature errors and its adept handling of error accumulation.
\begin{table}
  \centering
  \caption{Speedup ratio and average acceptance length $\tau$ on HumanEval, GSM8K, and Alpaca. T denotes temperature, V represents Vicuna, and LC stands for LLaMA2-Chat.}
  \resizebox{\columnwidth}{!}{
    \begin{tabular}{cccccccc}
    \toprule
          &       & \multicolumn{2}{c}{HumanEval} & \multicolumn{2}{c}{GSM8K} & \multicolumn{2}{c}{Alpaca} \\
    \midrule
          & Model & Speedup & $\tau$     & Speedup & $\tau$     & Speedup & $\tau$ \\
    \midrule
    \multirow{6}[2]{*}{T=0} & V 7B  & 3.33x & 4.29  & 3.01x & 4.00  & 2.79x & 3.86 \\
          & V13B  & 3.58x & 4.39  & 3.08x & 3.97  & 3.03x & 3.95 \\
          & V 33B & 3.67x & 4.28  & 3.25x & 3.94  & 2.97x & 3.61 \\
          & LC 7B & 3.17x & 4.24  & 2.91x & 3.82  & 2.78x & 3.71 \\
          & LC 13B & 3.76x & 4.52  & 3.20x & 4.03  & 3.01x & 3.83 \\
          & LC 70B & 3.52x & 4.42  & 3.03x & 3.93  & 2.97x & 3.77 \\
    \midrule
    \multirow{6}[2]{*}{T=1} & V 7B  & 2.39x & 3.43  & 2.34x & 3.29  & 2.21x & 3.30 \\
          & V13B  & 2.65x & 3.63  & 2.57x & 3.60  & 2.45x & 3.57 \\
          & V 33B & 2.76x & 3.62  & 2.77x & 3.60  & 2.52x & 3.32 \\
          & LC 7B & 2.61x & 3.79  & 2.40x & 3.52  & 2.29x & 3.33 \\
          & LC 13B & 2.89x & 3.78  & 2.82x & 3.67  & 2.66x & 3.55 \\
          & LC 70B & 2.92x & 3.76  & 2.74x & 3.58  & 2.65x & 3.47 \\
    \bottomrule
    \end{tabular}}
  \label{tab:other}%
\end{table}%

\begin{table}
  \centering
  \caption{Average acceptance length $\tau$ and acceptance rate $\alpha$ on MT-bench. T denotes temperature.}
  \resizebox{\columnwidth}{!}{
    \begin{tabular}{cccccccc}
    \toprule
          & Model & $\tau$     & $0\text{-}\alpha$    & $1\text{-}\alpha$    & $2\text{-}\alpha$    & $3\text{-}\alpha$    & $4\text{-}\alpha$ \\
    \midrule
    \multirow{6}[2]{*}{T=0} & Vicuna 7B  & 3.94  & 0.79  & 0.74  & 0.72  & 0.73  & 0.67 \\
          & Vicuna 13B  & 3.98  & 0.79  & 0.74  & 0.72  & 0.74  & 0.70 \\
          & Vicuna 33B & 3.68  & 0.74  & 0.69  & 0.67  & 0.67  & 0.66 \\
          & LLaMA2-Chat 7B & 3.62  & 0.76  & 0.69  & 0.67  & 0.68  & 0.68 \\
          & LLaMA2-Chat 13B & 3.90  & 0.77  & 0.69  & 0.69  & 0.70  & 0.71 \\
          & LLaMA2-Chat 70B & 3.81  & 0.75  & 0.69  & 0.65  & 0.64  & 0.64 \\
    \midrule
    \multirow{6}[2]{*}{T=1} & Vicuna 7B  & 3.17  & 0.71  & 0.68  & 0.66  & 0.66  & 0.65 \\
          & Vicuna 13B  & 3.20  & 0.73  & 0.68  & 0.68  & 0.67  & 0.69 \\
          & Vicuna 33B & 3.22  & 0.71  & 0.67  & 0.64  & 0.64  & 0.64 \\
          & LLaMA2-Chat 7B & 3.30  & 0.71  & 0.66  & 0.66  & 0.66  & 0.64 \\
          & LLaMA2-Chat 13B & 3.45  & 0.73  & 0.69  & 0.66  & 0.67  & 0.67 \\
          & LLaMA2-Chat 70B & 3.46  & 0.73  & 0.67  & 0.64  & 0.66  & 0.65 \\
    \bottomrule
    \end{tabular}}
  \label{tab:alpha}%
\end{table}%

Table \ref{tab:mix} reveals that EAGLE achieved a 1.5x speedup with the Mixtral 8x7B Instruct model. This modest acceleration, compared to models like LLaMA, is due to a shorter average acceptance length and the complexity of accelerating MoE models via speculative sampling. MoE models typically require reading the weights of only two experts per token during vanilla autoregressive decoding. However, during the verification phase of speculative sampling, processing multiple tokens may necessitate accessing the weights of more than two experts, contrasting with dense decoder-only models where all weights are read regardless of the number of tokens forwarded.
\begin{table}
  \centering
  \caption{Speedup ratio, average acceptance length $\tau$, and acceptance rate $\alpha$ on MT-bench at temperature=0. The target LLM is Mixtral 8x7B Instruct-v0.1.}
    \begin{tabular}{ccccccc}
    \toprule
    Speedup & $\tau$   & $0\text{-}\alpha$    & $1\text{-}\alpha$    & $2\text{-}\alpha$    & $3\text{-}\alpha$    & $4\text{-}\alpha$ \\
    \midrule
    1.50x & 3.25  & 0.67  & 0.62  & 0.61  & 0.64  & 0.63 \\
    \bottomrule
    \end{tabular}%
  \label{tab:mix}%
\end{table}%
\subsection{Case study: EAGLE + gpt-fast}
EAGLE is compatible with other acceleration technologies. We conducted experiments combining EAGLE with gpt-fast, which employs quantization and compilation for acceleration. As shown in Figure \ref{tab:gptfast}, by integrating EAGLE with gpt-fast, we increased the generation speed of LLaMA2-Chat 7B on a single RTX 3090 to 160.4 tokens/s.

\begin{table}
  \centering
  \caption{Generation speed of EAGLE combined with gpt-fast, evaluated on MT-bench with LLaMA2-Chat 7B at temperature=0.}
    \begin{tabular}{ccc}
    \toprule
    Precision & FP16  & int4 \\
    \midrule
    Vanilla (Huggingface) & 24.5 tokens/s & N/A \\
    gpt-fast & 55.1 tokens/s & 106.9 tokens/s \\
    EAGLE + gpt-fast & 100.2 tokens/s & 160.4 tokens/s \\
    \bottomrule
    \end{tabular}%
  \label{tab:gptfast}%
\end{table}%

\subsection{Ablation study}
\subsubsection{Tree attention}
\label{sec:abtree}
\begin{figure}[t]
\begin{center}
\centerline{\includegraphics[width=\columnwidth]{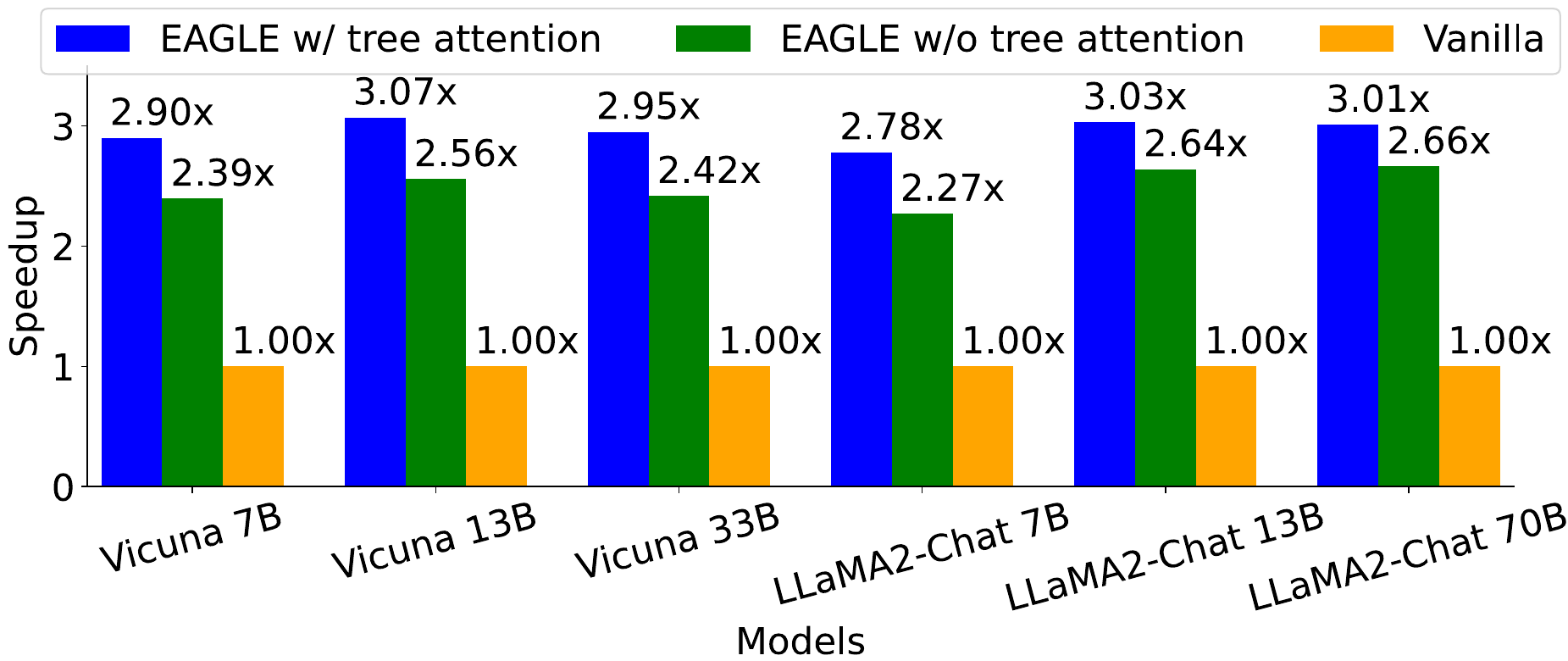}}
\caption{Speedup ratios of EAGLE with and without the use of tree attention. The evaluation dataset is MT-bench, with the temperature parameter set to 0.}
\label{aba-chain}
\end{center}
\end{figure}

EAGLE, similar to SpecInfer and Medusa, employs tree attention, where both the generation and validation of drafts are tree-structured. In contrast, methods like speculative sampling do not use tree attention, resulting in chain-structured draft generation and validation. Table \ref{tab:aba-chain} and Figure \ref{aba-chain} present comparative results indicating the impact of using tree attention. The implementation of tree draft and verification in EAGLE results in an approximate increase of 0.6-0.8 in the average acceptance length and about 0.3-0.5 in the speedup ratio. Compared to chain draft and verification, tree draft and verification do not increase the number of forward passes in the model (both the target LLM and the draft model), but they do increase the number of tokens processed per forward pass. Consequently, the improvement in the speedup ratio is less pronounced than the increase in average acceptance length. Notably, even without employing tree draft and verification, EAGLE demonstrates a significant acceleration effect, approximately in the range of 2.3x-2.7x.

\begin{table}[t]
  \centering
  \caption{Average acceptance length $\tau$ of EAGLE with and without the use of tree attention. The evaluation dataset is MT-bench, with the temperature parameter set to 0.}
  \resizebox{\columnwidth}{!}{
    \begin{tabular}{cccccc}
    \toprule
    \multicolumn{3}{c}{Vicuna} & \multicolumn{3}{c}{LLaMA2-Chat} \\
    \midrule
    Size  & Chain & Tree  & Size  & Chain & Tree  \\
    \midrule
    7B    & 3.20  & 3.94 (+0.74) & 7B    & 3.00  & 3.62 (+0.62) \\
    13B   & 3.23  & 3.98 (+0.75) & 13B   & 3.18  & 3.90 (+0.68) \\
    33B   & 2.97  & 3.68 (+0.71) & 70B   & 3.12  & 3.81 (+0.69) \\
    \bottomrule
    \end{tabular}}
  \label{tab:aba-chain}%
\end{table}%

\subsubsection{Inputs of draft models}
\label{sec:input}
\begin{figure*}[t]
\begin{center}
\includegraphics[width=\textwidth]{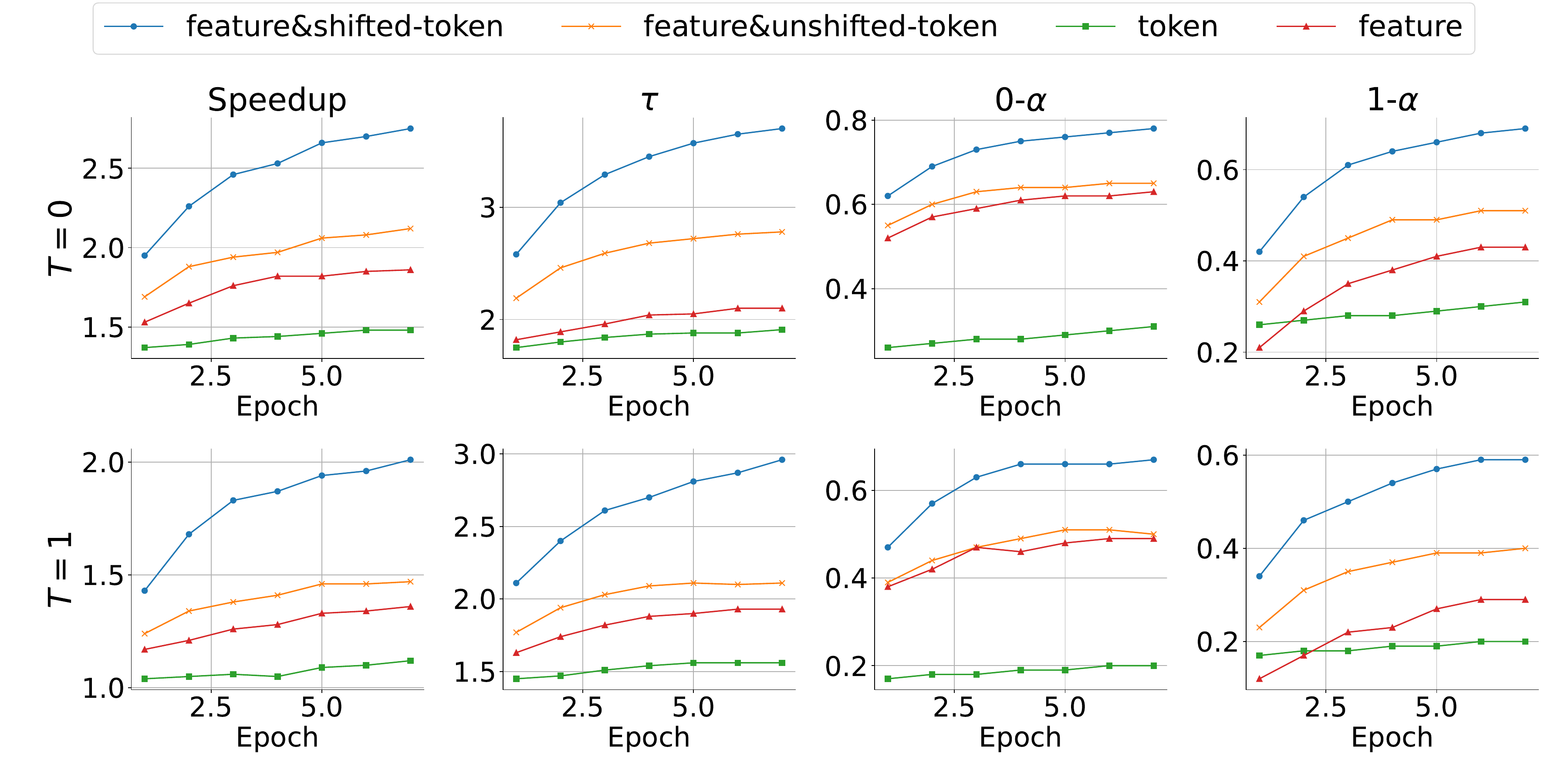}
\caption{Performance of draft models with varying inputs. The target LLM is Vicuna 7B, and the test dataset is MT-bench. Speed refers to the walltime speedup ratio, $\tau$ denotes the average acceptance length, $0\text{-}\alpha$ represents the acceptance rate with entirely precise inputs, $1\text{-}\alpha$ indicates the acceptance rate when the input includes one imprecise feature, and $T$ refers to the temperature.}
\label{fig:aba-input}
\end{center}
\vskip -0.1in
\end{figure*}

Compared to other speculative sampling-based methods, the key innovation of EAGLE lies in its utilization of features computed by the target LLM and the incorporation of sampling outcomes into the input of the draft model to address randomness. We conducted an ablation study on Vicuna 7B, assessing draft models with varying inputs. We tested four types of inputs: feature\&shifted-token (EAGLE), feature\&unshifted-token, token, and feature. Both feature\&shifted-token (EAGLE) and feature\&unshifted-token integrate semantic information at different levels. The distinction lies in the fact that feature\&shifted-token (EAGLE) inputs tokens advanced by one time step, equipping it to address randomness effectively. Apart from the use of a FC layer to reduce dimensionality for the feature\&token input, the structure of the draft model remains entirely consistent. Figure \ref{fig:aba-input} presents the experimental outcomes on the MT-bench with Vicuna 7B as the target LLM. Three observations can be drawn.
\begin{itemize}
    \item First, when the number of parameters of the draft model is limited, utilizing features yields slightly better results than tokens.
    \item Second, merging features and tokens modestly boosts performance, mainly as discrete, error-free tokens mitigate feature error accumulation, evident from the similar $0\text{-}\alpha$ of feature\&unshifted-token and feature-only draft models, with a significantly improved $1\text{-}\alpha$.
    \item Third, addressing the randomness inherent in the sampling process results in the most significant improvement. The feature\&shifted-token scheme, compared to feature\&unshifted-token, adds no complexity yet markedly enhances the draft model's capability by simply advancing the token by one time step, allowing the draft model to account for the randomness in sampling.
\end{itemize}

\subsubsection{Training data}
\label{sec:abadata}

EAGLE uses a fixed dataset for training, avoiding increased overhead from using the target LLM for generating training data. Ablation study (see Table \ref{tab:training data}) shows that data from the target LLM marginally improves performance, indicating EAGLE's low sensitivity to training data and justifying the fixed dataset approach for cost reduction.

\begin{table}
  \centering
  \caption{The speedup ratios and average acceptance length $\tau$ using different training datasets evaluated on the MT-bench, with the target LLM being LLaMA2-Chat 7B and the temperature set to 0. ``Fixed dataset'' refers to both questions and answers originating from the ShareGPT dataset. ``Data generated by target LLM'' denotes that while questions are sourced from the ShareGPT dataset, the answers are generated by the target LLM.}
    \begin{tabular}{ccc}
    \toprule
    Training data & Speedup & $\tau$ \\
    \midrule
    Fixed dataset & 2.78x & 3.62 \\
    Data  generated by target LLM & 2.88x & 3.75 \\
    \bottomrule
    \end{tabular}%
  \label{tab:training data}%
\end{table}%

\subsection{Batch size and throughput}
\label{sec:bsne1}

Inference in LLMs is memory-bound \cite{patterson2004latency,shazeer2019fast}, leaving GPU computational resources underutilized. The principle behind the speculative sampling-based approach in enhancing generation speed lies in more effectively utilizing GPU computational resources. As the batch size increases, the available computational capacity of the GPU decreases, leading to a reduction in the acceleration effect. In this section, we present experimental results for scenarios where the batch size exceeds 1. As demonstrated in Table \ref{tab:bs}, the speedup ratio diminishes with increasing batch size. When using Vicuna 7B as the target LLM, the speedup ratio at bs=4 is higher than at bs=3. This is attributed to the fact that, during the verification phase of EAGLE, the target LLM processes multiple tokens in a single forward pass, and the processing at bs=4 is faster than at bs=3. In contrast, with vanilla autoregressive decoding where the target LLM processes one token per forward pass, the speeds at bs=3 and bs=4 are nearly identical.

Although speculative sampling-based methods predominantly focus on latency, we also investigated EAGLE's throughput for batch size $>1$, another key metric for LLM systems. Compared to vanilla autoregressive decoding, EAGLE requires slightly more CUDA memory. For Vicuna 7B as the target LLM, operating under a memory constraint of a single RTX 3090 with 24G of CUDA memory, the maximum batch size (bs) for vanilla autoregressive decoding and EAGLE are 8 and 7, respectively. In the case of LLaMA2-Chat 70B, constrained by 4 A100 (40G) GPUs totaling 160G of CUDA memory, the maximum bs for vanilla autoregressive decoding and EAGLE are 5 and 4, respectively. All evaluations were conducted at FP16 precision. We calculated the throughput for different bs and selected the maximum value. Both vanilla autoregressive decoding and EAGLE achieve maximum throughput at their respective maximum bs. Tree attention consumes more computational resources. At bs=7, the computational resources are less abundant, making the non-use of tree attention more advantageous. As illustrated in Table \ref{tab:bs}, EAGLE achieves a 2x increase in throughput.

\begin{table}
  \centering
  \caption{Speedup ratios at different batch sizes and throughput of EAGLE. The evaluation dataset is MT-bench, with the temperature parameter set to 0.}
  \resizebox{\columnwidth}{!}{
    \begin{tabular}{cccccc}
    \toprule
    Batch size & 1     & 2     & 3     & 4     & Throughput \\
    \midrule
    Vicuna 7B & 2.90x & 2.87x & 2.65x & 2.76x & 1.97x \\
    LLaMA2-Chat 70B & 3.01x & 2.81x & 2.50x & 2.40x & 1.99x \\
    \bottomrule
    \end{tabular}}
  \label{tab:bs}%
\end{table}%

% \subsubsection{Throughput}

% The objective of the draft-verification approach is to reduce the generation latency and increase the generation speed of LLMs, rather than enhancing the overall throughput of LLM systems. Latency and throughput are two crucial metrics for LLMs, often presenting a trade-off. In our research, we have specifically investigated the throughput of EAGLE.

% cuda memory

\section{Related Work}

There has been considerable research into accelerating language models, involving techniques such as distillation \cite{hinton2015distilling}, quantization \cite{hubara2018quantized,shen2020q,kim2021bert,zadeh2020gobo,zafrir2019q8bert}, pruning \cite{gale2019state,sanh2020movement,kurtic2022optimal,voita2019analyzing}, and innovative network architecture designs \cite{gu2023mamba,wu2020lite}. These methods aim to reduce the latency per forward pass.

Similar to our approach are frameworks based on speculative sampling. Early works \cite{stern2018blockwise,sun2021instantaneous} accelerated greedy decoding, while speculative sampling \cite{leviathan2023fast,chen2023accelerating} extended it to non-greedy sampling, provably maintaining the original output distribution. Ensuring unchanged output distribution makes acceleration more challenging; many studies have explored lossy acceleration as a trade-off. For instance, DistillSpec \cite{zhou2023distillspec} modifies acceptance probabilities using a lenience function, BiLD \cite{kim2023speculative} accepts drafts if the distance metric from the target LLM distribution is below a certain threshold, and Medusa \cite{medusa} uses a minimum of a hard threshold and an entropy-dependent threshold for truncation. In contrast, EAGLE does not employ any relaxations and maintains the output distribution of the LLM unchanged.

The primary differences among speculative sampling-based methods manifest predominantly in the drafting phase. Speculative sampling \cite{leviathan2023fast,chen2023accelerating} utilizes a lower-parameter version of the target LLM as the draft model. Self-Speculative Decoding \cite{zhang2023draft} skips some layers of the target LLM during draft generation. SpecInfer \cite{miao2023specinfer} employs a set of small models to generate drafts in parallel. Cascade Speculative Drafting \cite{chen2023cascade} and Staged Speculative Decoding \cite{spector2023accelerating} cascade different overhead draft models. 
Online Speculative Decoding \cite{liu2023online} trains the draft model on a distribution of queries.
Methods \cite{hooper2023speed,fu2023lookahead,yang2023predictive} such as Medusa \cite{medusa} do not employ a separate target LLM; instead, they generate drafts by utilizing features or weights from the target LLM. REST \cite{he2023rest} generates drafts based on retrieval methods.
LLMA \cite{yang2023inference}, used for tasks like grammatical correction where input and output overlap, retrieves drafts directly from the input.

\section{Conclusion}
In this paper, we introduce EAGLE, an efficient framework for speculative sampling. EAGLE conducts the drafting process autoregressively at the more structured (second-to-top-layer) feature level and mitigates sampling uncertainty in predicting the next feature by incorporating tokens from one time step ahead. EAGLE is guaranteed to preserve the output distribution of the LLM while significantly enhancing generation speed. On MT-bench, EAGLE is 2.1x-3.8x faster than vanilla autoregressive decoding, 1.7x-2.1x faster than Lookahead, and 1.5x-1.6x faster than Medusa.

\textbf{Acknowledgements.} We acknowledge useful discussions with the Medusa's team leader Tianle Cai, the Lookahead's team leader Hao Zhang, the SpecTr's team leader Ziteng Sun, interactions with the gpt-fast team leaders Horace He and Soumith Chintala on X, and Yihan Wu.

\bibliography{example_paper}
\bibliographystyle{icml2024}

\newpage
\appendix

\onecolumn
\section{Implementation Details}
\subsection{Tree Structure}
\label{ap:tree}
\begin{figure*}[h]
    \centering
    \includegraphics[width=1\linewidth]{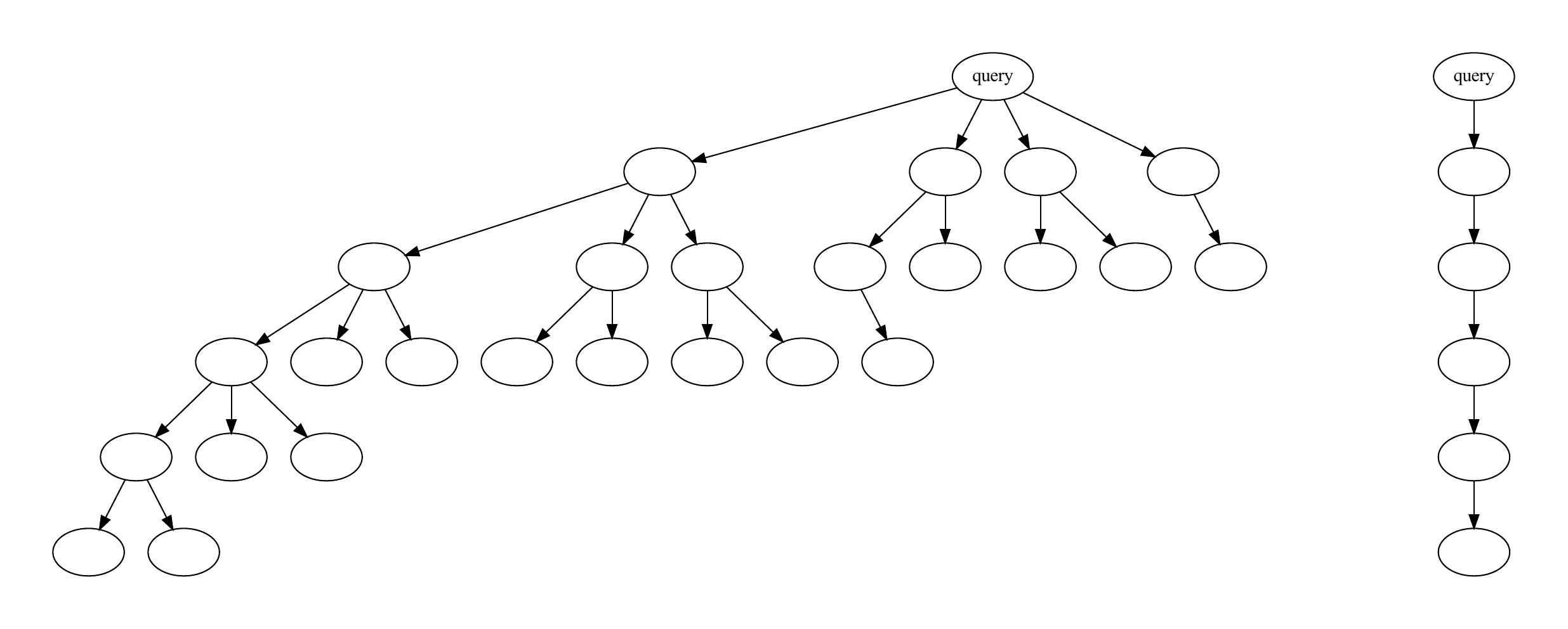}
    \caption{Structure of EAGLE's draft. The left side shows the draft structure when tree attention is employed, while the right side depicts the draft structure without the use of tree attention.}
    \label{fig:tree}
\end{figure*}

Utilizing tree attention, EAGLE generates a tree-structured draft. The left side of Figure \ref{fig:tree} illustrates the tree structure of the draft, while the right side depicts the corresponding chain-structured draft when tree attention is not used (as utilized in the ablation study detailed in Section \ref{sec:abtree}). In a greedy setting, we select the top $k$ tokens with the highest probabilities as child nodes. In a non-greedy setting, we sample $k$ tokens. The number of child nodes, $k$, can be inferred from Figure 9; for instance, $k=4$ at the root node. 
Regardless of employing a tree-structured or chain-structured draft, the draft model undergoes 5 forward passes during the draft phase.
During the verification phase, each token's probability is obtained through a single forward pass by the target LLM.

\textbf{Why do we use such a tree structure?} The choice of the tree structure, as depicted in Figure 9, was not rigorously optimized but rather based on intuition: branches of higher-probability tokens should be deeper and wider. For this paper, all models across all experiments utilized the draft structure shown in Figure 9. However, the optimal tree structure is likely context-dependent. For instance, as batch size increases and redundant computational resources decrease, a smaller tree might be preferable. Tuning the draft structure could potentially lead to improved performance.

\subsection{Multi-Round Speculative Sampling}
\label{ap:mss}
Unlike the chain-structured draft of speculative sampling, EAGLE employs a tree-structured draft, necessitating modifications to the sampling algorithm. The sampling algorithm A of speculative sampling can be briefly described as: if a token is accepted, it returns that token; otherwise, it samples a token from the adjusted distribution. For a tree-structured draft with $k$ candidate tokens, Multi-round speculative sampling recursively invokes algorithm A. Instead of directly sampling from the adjusted distribution after rejecting a token, Multi-round speculative sampling calls A again. If all tokens are rejected, it then directly samples from the adjusted distribution. The pseudocode for Multi-round speculative sampling is provided in Algorithm \ref{alg:mss}.

\begin{algorithm}
   \caption{Multi-round speculative sampling}
   \label{alg:mss}
\begin{algorithmic}
   \STATE {\bfseries Input:}  Target distribution $p$, samples $t_i$ and distributions $\hat{p_i}$ for each $i$ from 1 to $k$, where $t_i$ is sampled from $\hat{p_i}$,
   \STATE {\bfseries Output:} a sample $x \sim p$ ;
   \STATE  $i \leftarrow 1$
   \FOR {$i \leq k$}
   \STATE $r \leftarrow U(0,1)$ 
   \IF{$r<{p(t_i)}/{\hat{p_i}(t_i)}$}
   \STATE {\bfseries Return} $t_i$
   \ENDIF
   \STATE $p \leftarrow norm(max(0,p(t_i)-\hat{p_i}(t_i)))$ 
   \STATE $i \leftarrow i+1$
   \ENDFOR
\STATE Sample $t \sim p$
\STATE {\bfseries Return} $t$
\end{algorithmic}
\end{algorithm}

\section{Detailed experimental results}
\label{apex}

Table \ref{tab:a} displays the speedup ratio, average acceptance length $\tau$ and acceptance rate $\alpha$ of EAGLE on HumanEval, GSM8K, and Alpaca datasets.

\begin{table*}
  \centering
  \caption{Speedup ratio, average acceptance length $\tau$ and acceptance rate $\alpha$ on HumanEval, GSM8K, and Alpaca at temperature = 0.}
    \begin{tabular}{ccccccccc}
    \toprule
    Dataset & Model & Speedup & $\tau$     & $0\text{-}\alpha$    & $1\text{-}\alpha$    & $2\text{-}\alpha$    & $3\text{-}\alpha$    & $4\text{-}\alpha$ \\
    \midrule
    \multirow{6}[2]{*}{HumanEval} & Vicuna 7B & 3.33x & 4.29  & 0.82  & 0.77  & 0.72  & 0.69  & 0.71 \\
          & Vicuna13B & 3.58x & 4.39  & 0.85  & 0.78  & 0.74  & 0.72  & 0.73 \\
          & Vicuna 33B & 3.67x & 4.28  & 0.83  & 0.77  & 0.74  & 0.70  & 0.70 \\
          & LLaMA2-Chat 7B & 3.17x & 4.24  & 0.81  & 0.76  & 0.73  & 0.74  & 0.72 \\
          & LLaMA2-Chat 13B & 3.76x & 4.52  & 0.85  & 0.80  & 0.78  & 0.76  & 0.75 \\
          & LLaMA2-Chat 70B & 3.52x & 4.42  & 0.84  & 0.79  & 0.75  & 0.73  & 0.74 \\
    \midrule
    \multirow{6}[2]{*}{GSM8K} & Vicuna 7B & 3.01x & 4.00  & 0.79  & 0.71  & 0.70  & 0.71  & 0.70 \\
          & Vicuna13B & 3.08x & 3.97  & 0.79  & 0.71  & 0.67  & 0.68  & 0.64 \\
          & Vicuna 33B & 3.25x & 3.94  & 0.79  & 0.71  & 0.67  & 0.67  & 0.67 \\
          & LLaMA2-Chat 7B & 2.91x & 3.82  & 0.75  & 0.69  & 0.64  & 0.65  & 0.63 \\
          & LLaMA2-Chat 13B & 3.20x & 4.03  & 0.80  & 0.70  & 0.70  & 0.68  & 0.66 \\
          & LLaMA2-Chat 70B & 3.03x & 3.93  & 0.77  & 0.71  & 0.66  & 0.64  & 0.60 \\
    \midrule
    \multirow{6}[2]{*}{Alpaca} & Vicuna 7B & 2.79x & 3.86  & 0.74  & 0.68  & 0.66  & 0.66  & 0.67 \\
          & Vicuna13B & 3.03x & 3.95  & 0.72  & 0.67  & 0.64  & 0.63  & 0.64 \\
          & Vicuna 33B & 2.97x & 3.61  & 0.70  & 0.64  & 0.64  & 0.63  & 0.64 \\
          & LLaMA2-Chat 7B & 2.78x & 3.71  & 0.73  & 0.66  & 0.62  & 0.64  & 0.62 \\
          & LLaMA2-Chat 13B & 3.01x & 3.83  & 0.75  & 0.67  & 0.64  & 0.63  & 0.63 \\
          & LLaMA2-Chat 70B & 2.97x & 3.77  & 0.76  & 0.68  & 0.65  & 0.61  & 0.62 \\
    \bottomrule
    \end{tabular}%
  \label{tab:a}%
\end{table*}%

\end{document}